\documentclass{article}

\usepackage[preprint, nonatbib]{neurips_2021}

\usepackage[utf8]{inputenc} 
\usepackage[T1]{fontenc}    
\usepackage{hyperref}       
\usepackage{url}            
\usepackage{booktabs}       
\usepackage{amsfonts}       
\usepackage{nicefrac}       
\usepackage{microtype}      
\usepackage{xcolor}         
\usepackage{amsmath}
\usepackage{subfigure}
\usepackage{graphicx}
\usepackage[backend=bibtex]{biblatex}
\newtheorem{lemma}{Lemma}
\newtheorem{thm}{Theorem}

\title{Barcode Method for Generative Model Evaluation driven by Topological Data Analysis}

\author{%
  Ryoungwoo Jang \\
  VUNO Incorporation \\
  Seoul, South Korea \\
  \texttt{ryoungwoo.jang@vuno.co} \\
  \And
  Minjee Kim \\
  University of Ulsan, College of Medicine\\
  Asan Medical Center\\
  Seoul, South Korea \\
  \texttt{minjeekim00@gmail.com} \\
  \AND
  Da-in Eun \\
  University of Ulsan, College of Medicine\\
  Asan Medical Center\\
  School of Medicine, Kyunghee University\\
  Seoul, South Korea \\
  \texttt{eundain94@gmail.com} \\
  \And
  Kyungjin Cho \\
  University of Ulsan, College of Medicine\\
  Asan Medical Center\\
  Seoul, South Korea \\
  \texttt{kjcho.amc@gmail.com} \\
  \And
  Jiyeon Seo \\
  University of Ulsan, College of Medicine\\
  Asan Medical Center\\
  Seoul, South Korea \\
  \texttt{jiyeon.amc@gmail.com} \\
  \And
  Namkug Kim\thanks{Correspondence to Namkug Kim} \\
  University of Ulsan, College of Medicine\\
  Asan Medical Center\\
  Seoul, South Korea \\
  \texttt{namkugkim@gmail.com}
}

\begin{document}

\maketitle

\begin{abstract}
    Evaluating the performance of generative models in image synthesis is a challenging task. Although the Fr\'echet Inception Distance is a widely accepted evaluation metric, it integrates different aspects (e.g., fidelity and diversity) of synthesized images into a single score and assumes the normality of embedded vectors. Recent methods such as precision-and-recall and its variants such as density-and-coverage have been developed to separate fidelity and diversity based on $k$-nearest neighborhood methods. In this study, we propose an algorithm named \textbf{barcode}, which is inspired by the topological data analysis and is almost free of assumption and hyperparameter selections. In extensive experiments on real-world datasets as well as theoretical approach on high-dimensional normal samples, it was found that the `usual' normality assumption of embedded vectors has several drawbacks. The experimental results demonstrate that \textbf{barcode} outperforms other methods in evaluating fidelity and diversity of GAN outputs. Official codes can be found in \url{https://github.com/minjeekim00/Barcode}
\end{abstract}

\section{Introduction}

Since the introduction of generative adversarial networks (GAN) \cite{goodfellow2014generative}, generative models have encountered a new era. New architectures were developed for high-quality image generation \cite{radford2015unsupervised, karras2017progressive, brock2018large}, image synthesis \cite{karras2019style, karras2020analyzing}, image-to-image translation \cite{Choi2018StarGANUG, Choi2020StarGANVD, Zhu2017UnpairedIT}, etc. In addition, a training method that can generate high-resolution images with limited datasets \cite{karras2020training} has been introduced. Although these GANs work well for generating photo-realistic images at high resolution, quantitative evaluation of performance for these models is still difficult.

Many had tried to develop a evaluation metric for generative models. Inception score (IS) was suggested by Salimans \textit{et al} \cite{salimans2016improved}, of which their basic idea is based on entropy in information theory. The IS uses a pre-trained network, Inception-V3 \cite{szegedy2016rethinking}, to classify the real image and the synthetic images. Formally, the IS is formulated as follows:
\begin{align*}
\exp(\mathbb{E}_x\text{KL}(p(y|\textbf{x})||p(y)))=\exp\bigg(\mathbb{E}_x\mathbb{E}_{p(y|\textbf{x})}\bigg[\log\bigg(\frac{p(y|\textbf{x})}{p(y)}\bigg)\bigg]\bigg)
\end{align*}
Based on this idea, Fr\'echet Inception distance (FID) had been suggested \cite{heusel2017gans}. 
If $m$ is the real mean vector, $m_w$ is the synthetic mean vector, $\Sigma$ is the covariance matrix of the real dataset, and $\Sigma_w$ is the covariance matrix of the synthetic dataset, then the FID is defined as:
\begin{align*}\label{eq:eq1}
d^2((m,\mathcal{C}),(m_w,\mathcal{C}_w))=||m-m_w||_2^2+\text{Tr}(\Sigma+\Sigma_w-2(\Sigma\Sigma_w)^{1/2})
\end{align*}
FID assumed that coding units, also known as embedding vectors, follow the Gaussian distribution, which is unrealistic (this argument is discussed in section \ref{theoretical_approach}.). It is natural for GANs to be trained on real images and generate fake images based on Gaussian distribution. However, GANs do work as change of variables. That is, GANs change multivariate Gaussian distribution with relatively low-dimension (usually this is 512-dimensional random vector) to another high-dimensional image distribution, ranging from 512$\times$512 to 1024$\times$1024, which is intractable. Therefore, Gaussian distribution is not adequate distribution for intractable image distribution. Obviously, Fr\'echet distance can be expressed in closed form only for tractable, easy distributions. This seems FID score is not adequate for measuring two embedding vector sets in high dimensions. To detour this problem, Sajjadi \textit{et al} \cite{sajjadi2018assessing} developed a new concept named precision and recall (P\&R). Precision is a concept that how much generator can generate part of real images and recall is a concept of vice versa. At first glance, this does not seem computationally feasible due to the fact that we can not explicitly express the support of two distributions. Therefore the authors therefore suggested a feasible algorithm based on a resolution parameter $m$ and a calculation method based on $\lambda$ from 0 to $\infty$. To improve this P\&R method, Kynk\"a\"anniemi \textit{et al} \cite{kynkaanniemi2019improved} proposed an improved P\&R (iP\&R) algorithm based on $k$-nearest neighborhood ($k$NN). This iP\&R algorithm has some drawbacks. The most significant drawback of the iP\&R algorithm is that it is vulnerable to outliers. This is explained in Naeem \textit{et al} \cite{naeem2020reliable}. Therefore, in this paper, the authors suggested an outlier-robust algorithm with the introduction of the concepts, precision, recall, density and coverage (PRDC).

In iP\&R, precision and recall are defined as:
\begin{equation*}
\text{precision}=\frac{1}{M}\sum_{j=1}^M1_{Y_j\in\text{manifold}(X_1,\cdots,X_N)}
\end{equation*}
\begin{equation*}
\text{recall}=\frac{1}{N}\sum_{j=1}^N1_{Y_j\in\text{manifold}(Y_1,\cdots,Y_N)}
\end{equation*}

And in PRDC the density and coverage is defined as 
\begin{equation*}
 \text{dentisy}=\frac{1}{kM}\sum_{j=1}^M\sum_{i=1}^N1_{Y_j\in B(X_i, \text{NND}_k(X_i))}
\end{equation*}
\begin{equation*}
\text{coverage}=\frac{1}{N}\sum_{i=1}^N1_{\exists j\text{ s.t. }Y_j\in B(X_i,\text{NND}_k(X_i))}
\end{equation*}
where $\text{NND}_k$ uses the $k$NN method. In general, $k$ is set from 3 to 5. Since $k$NN has a hyperparameter $k$, the decision of how to select is another concern.

In this paper, we introduce a hyperparameter-free, topology-based discrete GAN evaluation metric that can be applied for GAN-generated images. Our metric is inspired by topological data analysis (TDA) \cite{carlsson2005persistence} on bipartite graph and uses the notion of barcode in persistent homology. With our proposed method, one can calculate the discrepancy between two distributions as well as how much GAN generates realistic outputs.

\section{Basic concepts}

\subsection{\v Cech complex}
Let $X$ be a metrizable topological space equipped with a metric $d$. Let $V\subset X$ be a finite set of points. Then $\sigma\subseteq V$ is an element of \textbf{\v Cech complex} $C_{\epsilon}(V)$, $\sigma\in C_{\epsilon}(V)$ if there are balls with radius $\epsilon/2$ and centered at each point of $\sigma$ that have non-empty intersection. If $|\sigma|=k$, then $\sigma$ is called a $k$-simplicial complex.
In our proposed method, we only use 1-simplicial complexes which are just line segments whose length is less than $\epsilon$. Note that the 1-simplicial complexes in {\v Cech complex} coincide with that of {\bf Vietoris-Rips} complexes.\\
Although we had supposed $V$ is a finite set, it is computationally expensive to calculate and record all intersections to get \v Cech complex (or Vietoris-Rips complex). Therefore, we detour this problem as follows: First, we record all distances between $V_P$, the real samples and $V_Q$, the synthetic samples as:
\begin{align}\label{eq1}
\begin{split}
d(V_P, V_Q) = \{d(v_P, v_Q):v_P\in V_P, v_Q\in V_Q, d:V_P\times V_Q\to\mathbb{R}^+\cup \{0\}\}
\end{split}
\end{align}
This set $d(V_P, V_Q)$ allows overlap, which means same value can be included more than twice. This is computationally cheap for the reason that we only use 1-simplicial complex, the lines with finite-length.

\subsection{Barcode}
After calculating Vietoris-Rips complex and distances between points, we introduce \textbf{barcode} which is a notion used in TDA. Although our barcode is not exactly same concept of TDA - which uses persistant homology of algebraic topology - we analogously define barcode as following manner:
\begin{enumerate}
\item First, we have distances between points of partitions in bipartite graph, $V_P$ and $V_Q$ for all combinations.
\item Second, we find the number of combinations that have a distance smaller than $\lambda$. Here, $\lambda$ ranges from 0 to $\max\{d(v_P, v_Q):v_P\in V_P, v_Q\in V_Q\}$
\item Third, we plot the number of combinations in $y$-axis depending on $\lambda$. This plot looks like a \textbf{barcode} image, from which the name of our proposed algorithm comes. Example of this barcode plot is shown in Figure (\ref{barcode_ex})
\item Fourth, we normalize $y$-axis with total number of combinations and normalize $x$-axis with $\max\{d(v_P, v_Q):v_P\in V_P, v_Q\in V_Q\}$. The set of normalized distances are denoted as $\tilde{d}(V_P,V_Q)$.This gives us a cumulative distribution function (CDF)-like curve plotted in $[0,1]\times [0,1]$.
\item Fifth, we calculate the area above the CDF-like curve (AUbC). This AUbC is called \textbf{fidelity}.
\item Sixth, we calculate standard deviation of this CDF-lke curve. This standard deviation is called \textbf{diversity}.
\end{enumerate}
Note that our proposed barcode method uses 1-dimensional Vietoris-Rips complex. It is more intuitive to call our proposed metric as 1-barcode, yet in this paper we just use the word, barcode for 1-barcode.

\subsection{Equivalent definition of fidelity}
We had defined fidelity as AUbC of the CDF-like curve. However, this can be computed faster than just implementing original definition. In this section, we calculate barcode and show that AUbC is equivalent to the mean of normalized CDF-like curve of barcode. After normalization, fidelity is defined as follows:
\begin{equation}
    \text{fidelity} = \int_{0}^{1}\big(1-F(x)\big)dx
\end{equation}
If we consider $F(x)$ to be CDF of some random variable, $X$, this is equivalent to
\begin{align}
    \text{fidelity}     &=\int_0^1P(X\geq x)dx\label{change2}\\
    &=\mathbb{E}[X]\nonumber
\end{align}

\subsection{Extrinsic, intrinsic and relative fidelity}
We had defined fidelity, but fidelity itself is a subjective value, not an absolute value. Therefore, we define \textbf{relative fidelity}, which means how much generation models soundly generate synthetic images relative to the real ones. Obviously, it is natural to define relative fidelity as quotient of fidelity between real and synthetic images. Relative fidelity $\phi(P,Q)$ is defined as:
\begin{equation}\label{relative_fidelity}
\phi(P,Q) = \frac{\text{f}(P,Q)}{\text{f}(P,P)}
\end{equation}
where $\text{f}(\cdot,\ast)$ is fidelity between $\cdot$ and $\ast$. If we say fidelity without any adjective, we imply relative fidelity, $\phi(P,Q)$. We say $\text{f}(\cdot, \ast)$ as \textbf{extrinsic fidelity} when $\cdot\neq\ast$, and \textbf{intrinsic fidelity} when $\cdot=\ast$. Also, note that $P$ denotes real data.

\subsection{Extrinsic, intrinsic and relative diversity}
In this section, we will define intrinsic diversity and extrinsic diversity. Mutual diversity of two sets $P$ and $Q$ is defined as follows: Let $d(V_P, V_Q)$ be the set of triples in Equation \ref{eq1}. Then, mutual diversity $\delta(P,Q)$ between $P$ and $Q$ is defined as:
\begin{equation}
    \delta(P,Q) = \sqrt{\text{Var}[\tilde{d}(V_P,V_Q)]}
\end{equation}
Note that $\tilde{d}$ denotes normalized distances. \textbf{Extrinsic diversity} is just $\delta(P,Q)$. Note that $\delta$ is symmetric: $\delta(P,Q)=\delta(Q,P)$. Although we had defined fidelity and diversity, these two measures may not be able to differentiate the undesired situation. Consider Figure (\ref{fig2}). These two situations show similar fidelity and diversity. It seems that there is no way to distinguish between two different situations. However, this can be overcome with the concept, \textbf{intrinsic diversity}. Intrinsic diversity is defined as $\delta(P,P)$ and $\delta(Q,Q)$. As we had recorded every combination of two finite samples, $\delta(P,P)$ and $\delta(Q,Q)$ are not zeros unless they collapse into one point. Therefore, by measuring intrinsic diversity of red sets, we can conclude that left situation of Figure (\ref{fig2}) has a small intrinsic diversity compared to that of right situation of Figure (\ref{fig2}). Due to the fact that relative diversity is calculated from two intrinsic diversities ($\delta(P,P)$ and $\delta(Q,Q)$), we calculated relative diversity of two situations in Figure (\ref{fig2}).

Furthermore, by calculating 
\begin{equation}\label{relative_diversity}
\rho(P,Q) = \frac{\delta(P,Q)}{\sqrt{\delta(P,P)}\sqrt{\delta(Q,Q)}}
\end{equation}
we can get \textbf{relative diversity}, which can be interpreted as how much capacity that synthetic sample distribution ($Q$) has compared to the real sample distribution ($P$). If we say diversity without any adjective, we imply relative diversity, $\rho(P,Q)$. However, there is a minor issue. $\delta(P,P)$ and $\delta(Q,Q)$ are calculated with zero distances, from the fact that $d(V_P,V_P)$ and $d(V_Q,V_Q)$ contain distance of the same points. As zero distances affect standard deviation of $d(V_P,V_P)$ and $d(V_Q,V_Q)$, we excluded zero distance that comes from the same points. Note that zero distances that come from different points are not excluded.

\subsection{Range of fidelity, diversity}
We had defined fidelity, diversity for two distribution $P$ and $Q$. Clearly, mutual fidelity ranges from 0 to 1 due to normalization. It is less obvious, yet mutual diversity is bounded between 0 and 1/3 as well.
\begin{align*}
    \delta(P,Q)^2 &= \text{Var}[\tilde{d}(V_P,V_Q)] = \text{Var}[X]\\
    &= \int_0^1(x-\mu)^2dx\\
    &=\int_0^1 x^2dx - \mu^2\\
    &\leq\int_0^1x^2dx=\frac{1}{3}
\end{align*}

\subsection{Dimension reduction technique}\label{explainability}

Usually, high-dimensional data does not follow our intuition. For example, in high dimension, most random vectors have similar distance and random vectors sampled from Gaussian distribution are nearly orthogonal (See Lemma \ref{Lemma1}.). Therefore, dimension reduction on high-dimensional data is inevitable. Suppose we are given $N$ embedding vectors from distribution $P$ and $M$ embedding vectors from distribution $Q$ with both dimension $D$. In other words, we have $(N,D)$ shaped vectors from distribution $P$ and $(M,D)$ shaped vectors from distribution $Q$. $D>1000$ in most cases, thus we need to reduce dimension to embed embedding vectors in low dimension. With this purpose in mind, we use singular value decomposition (SVD) for dimension reduction. For example, if we have $N$ vectors in $\mathbb{R}^{D}$ space, namely $E\in\mathbb{E}^{N\times D}$, we use following decomposition:
\begin{equation}
E=U^T\Sigma V
\end{equation}
where $U$ and $V$ are the orthogonal matrices in the orthogonal group, $O(\mathbb{R}^N)$, $O(\mathbb{R}^D)$ respectively. We will use singular values in $\Sigma$ matrix. If we intend to reduce dimension into $D'<D$, we select the largest $\frac{D'}{D}$ proportion of singular values and calculate the ratio for singular values summation. Explainability is defined as 
\begin{equation}
\text{Explainability} = \frac{\lambda_1^2+\lambda_2^2+\cdots+\lambda_{D'}^2}{\lambda_1^2+\lambda_2^2+\cdots+\lambda_D^2}
\end{equation}
where $\lambda_i$ are singular values and $\lambda_1\geq\lambda_2\geq\cdots\geq\lambda_D$

\subsection{Outlier removal}\label{section_outlier_removal}
In some cases, one might want to remove outliers for some reason. For this case, we suggest an outlier removal algorithm based on distance sorting. Outliers are usually apart from the original distribution, therefore measuring distances between two distributions tells us that the outliers may locate in extreme positions. Therefore, we suggest sort distance and remove outliers that have too small distances or too large distances or both.

\section{Theoretical Approach and Experiments}\label{experiments}

In this section, we analyzed normality assumptions with theoretical approach and experiment various metrics on virtual and real dataset on several models, circumstances. 
\subsection{Theoretical and experiments on normal distribution}\label{theoretical_approach}
\subsubsection{Theoretical approach on virtual dataset}\label{section31}

\begin{figure}[ht]
\vskip 0.2in
\begin{center}
\centerline{\includegraphics[width=80mm]{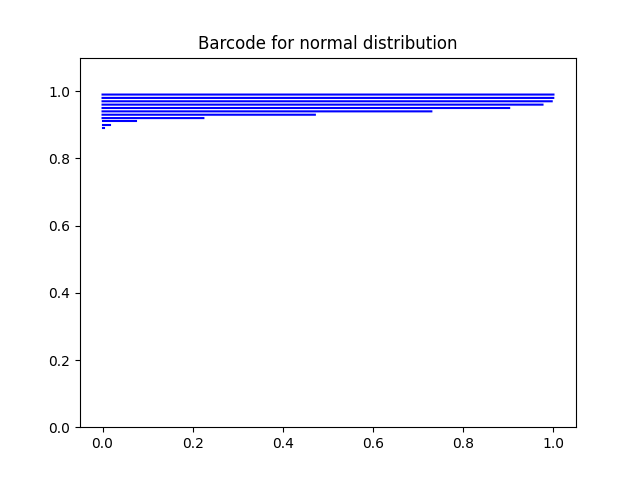}}
\caption{Barcode plot from two independent 2048 dimensional random normal distribution-sampled vectors with mean 0, covariance $\delta_{ij}$. Fidelity was 0.063, diversity was 1.002. }
\label{normal_barcode}
\end{center}
\vskip -0.2in
\end{figure}

We performed experiments on virtual dataset, which is sampled from high-dimensional normal distribution, and assert that normal assumption in models, such as FID is not a valid assumption. Our experimental scheme on virtual dataset with barcode measurement is:
\begin{enumerate}
\item We generated 10,000 samples from 2048 dimensional normal distribution which has independence between each dimension. That is, the covariance matrix of a 2048-dimensional normal distribution is 2048 identity matrices, $I_{2048}$.
\item We calculated barcode for this 2048-dimensional normal distribution and showed normal hypothesis is not valid compared to the experiments of section \ref{section32}. Here, explainability (see section \ref{explainability}) was set to be 1, implying we used the full dimension.
\item We also mathematically calculated properties of high-dimensional multivariate normal distribution and validated our experimental results with mathematical facts. As one can see the discrepancy of experimental results between virtual dataset and real datasets in section \ref{section32}, we claim that normal hypothesis in metrics (e.g. FID) may distort fundamental properties of real-world dataset.
\end{enumerate}
As mentioned above, we generated two 2048-dimensional vectors independently sampled from the normal distribution with the mean of 0, the covariance between $i$-th and $j$-th feature was $\delta_{ij}$, where $\delta_{ij}$ denotes Kronecker delta. We used normal function in numpy package random module with python during sampling. After then, barcode plot was plotted and calculated fidelity and diversity from this distribution. Here, explainability of the experiment was set to be 1, implying that we had used the full of 2048 dimension (Figure (\ref{normal_barcode})). In our experiment, extrinsic fidelity was 0.063, relative diversity was 1.002, showing low fidelity and high diversity.

In mathematical viewpoint, this can be explained with two perspectives. First is the perspective of Law of Large Numbers (LLN). Let $\textbf{x}$ and $\textbf{y}$ be two samples from $d$-dimensional random normal distribution with covariance being the $d$-dimensional identity matrix. Then, we define another variable $\textbf{z}$, namely $\textbf{z}=||\textbf{x}-\textbf{y}||_2^2$. Therefore, $i$-th element of $\textbf{z}$, which is $z_i$ is defined as $z_i=(x_i-y_i)^2$. Then, $z_i$s are all independent. Regardless of what distribution $z_i$ follows, LLN tells us that
\begin{equation}
P\bigg(\bigg|\frac{z_1+z_2+\cdots+z_d}{d}-\mathbb{E}[z_1]\bigg|\geq\epsilon\bigg)\leq\frac{\text{Var}(z_1)}{d\epsilon^2}
\end{equation}
due to the fact that $z_i$s are independent and identically distributed. \cite{blum2016foundations} Therefore, what LLN tells us is that as $d$ gets larger, variance of distance becomes smaller proportional to $1/d$. Therefore, random vectors sampled from high-dimension $d$ situation will have almost same distance with high probability. \\
In second perspective, this can be proved in Gaussian Annulus Theorem (GAT). Before applying GAT, we need following lemma. \cite{blum2016foundations}
\begin{lemma}[Near Orthogonality]\label{Lemma1}
Consider drawing $n$ points $\textbf{x}_1, \textbf{x}_2, \cdots, \textbf{x}_n$ from random unit ball. Then, with probability $1-O(1/n)$,
\begin{itemize}
    \item $|\textbf{x}_i|\geq1-\frac{2\ln n}{d}$ for $i=1,\cdots,n$
    \item $|\textbf{x}_i\cdot \textbf{x}_j|\leq\frac{\sqrt{6\ln n}}{\sqrt{d-1}}$ for $1\leq i\neq j\leq n$
\end{itemize}
\end{lemma}
GAT says that for a $d$-dimensional spherical normal with variance 1 for all variate, for any $\beta\leq\sqrt{d}$, all but $3e^{-c\beta^2}$ of the probability mass lies within the annulus $\sqrt{d}-\beta\leq|\textbf{x}|\leq\sqrt{d}+\beta$ \cite{blum2016foundations}. 
Therefore, if we sample two random vectors from high-dimensional normal distribution, most of the cases are sampled within some ``annulus" and they become orthogonal with high probability which is the result of Lemma \ref{Lemma1}, therefore distances between these two vectors will have same value with high probability.\\
To summarize, we get the following two results:
\begin{thm}\label{thm1}
Let $\textbf{x}_1, \textbf{x}_2, \cdots, \textbf{x}_n$ be $n$-randomly sampled $d$-dimensional vectors from normal distribution with mean 0, covariance matrix being identity matrix. Then following are asymptotically true:
\begin{enumerate}
\item As $d$ gets larger, $|\textbf{x}_i-\textbf{x}_j|$ becomes similar for all combinations of $1\leq i\neq j\leq n$.
\item As $d$ gets larger, $\textbf{x}_i\cdot\textbf{x}_j\approx 0$ for all combinations of $1\leq i\neq j\leq n$.
\end{enumerate}
\end{thm}

Experimental result of this proof is shown in Figure (\ref{normal_barcode}). As shown in Figure (\ref{normal_barcode}), sampled data from normal distribution shows that most of the distances are located in similar range, between 0.85 to 1.0 ratio. This is exact contradiction to our intuition. Therefore, Gaussian (normal) assumptions of metrics such as FID may be too strong assumptions to be applied to the real-world datasets.

\subsubsection{Experiments on virtual dataset}\label{section312}
We had analyzed normal distribution experiments following PRDC paper \cite{naeem2020reliable}. This is discussed in Appendix \ref{appendix}.

\subsection{Experiments on real datasets }\label{section32}



\subsubsection{Facial Dataset}
We had analyzed PRDC and our proposed barcode method for facial datasets, namely CelebA-HQ and FFHQ datasets. This is discussed in Appendix \ref{appendixc}

\subsubsection{Brain CT}
\begin{figure}[h]
\vskip 0.2in
\begin{center}
\centerline{\includegraphics[width=100mm]{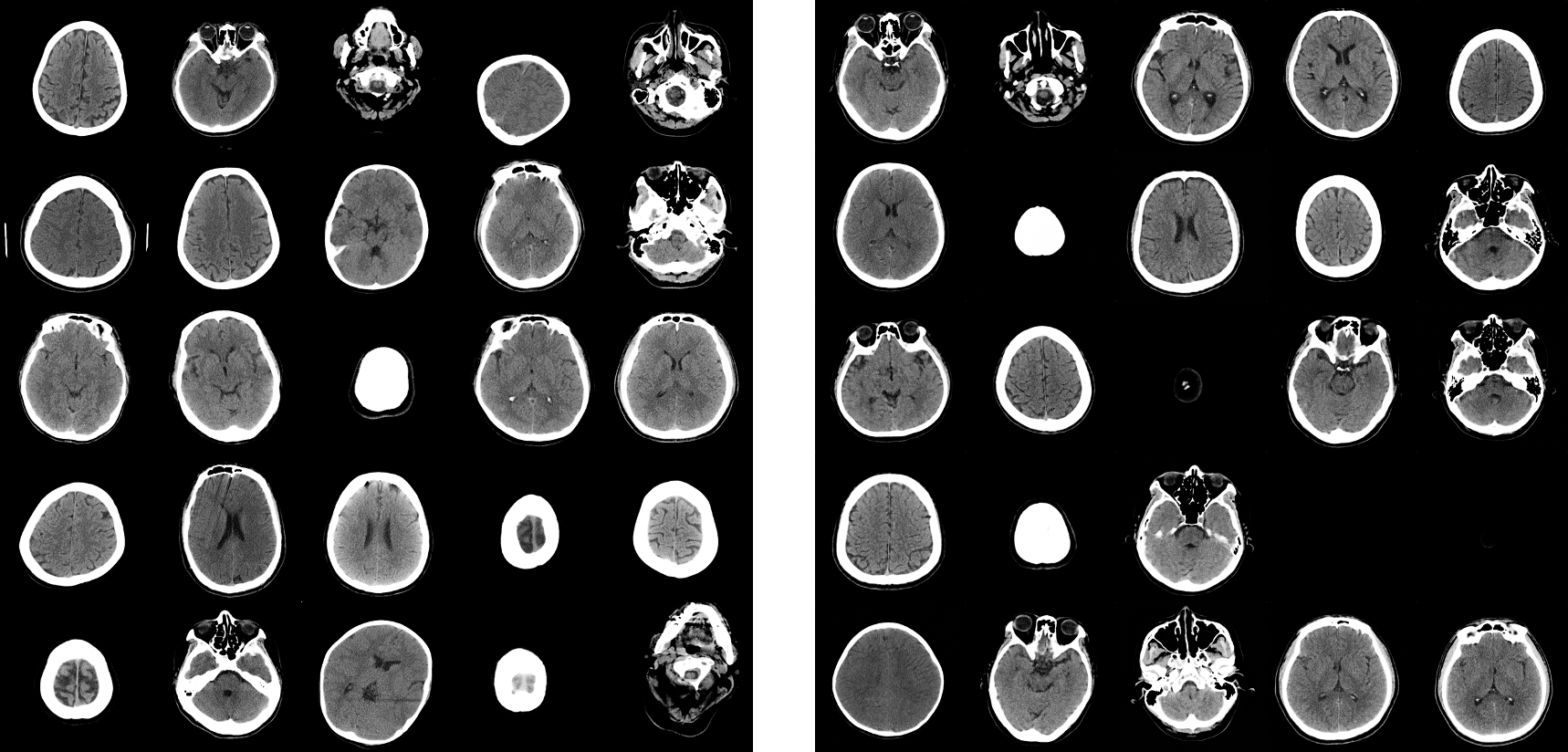}}
\caption{(Left) Randomly sampled 25 real brain CT images. (Right) Randomly synthetic 25 brain CT images with StyleGAN2 model. Original images have 3 channels yet in this figure we converted 3-channel into grayscale image for visual convenience.}
\label{brainct_ex}
\end{center}
\vskip -0.2in
\end{figure}

We collected 1,111,456 brain computed tomography (CT) images from 34,080 subjects at a single referral hospital. Original brain CT images are higher than 8-bit, usually 12-bit images, i.e. pixel values range from -1024 to 3071. Because 12-bit images contain too much information and 8-bit monitors can not contain high bit images without loss of pixel information, we used a windowing process that highlights a pixel range of interest. We windowed brain CT images into 3 windows that radiologists usually use in clinical settings on brain CT and used the 3 window as red, green, blue channels. All images were 512$\times$512 size, and set to be 3 channel as mentioned above. \\
We used StyleGAN2 for brain CT image generation. After the training process, we generated high-quality brain CT images synthetically. A radiologist over 10-years-experience could not notice the difference between real and synthetic images. Examples of sampling of real and synthetic images are shown in Figure (\ref{brainct_ex}).

\begin{figure}[h]
\vskip 0.2in
\begin{center}
\centerline{\includegraphics[width=100mm]{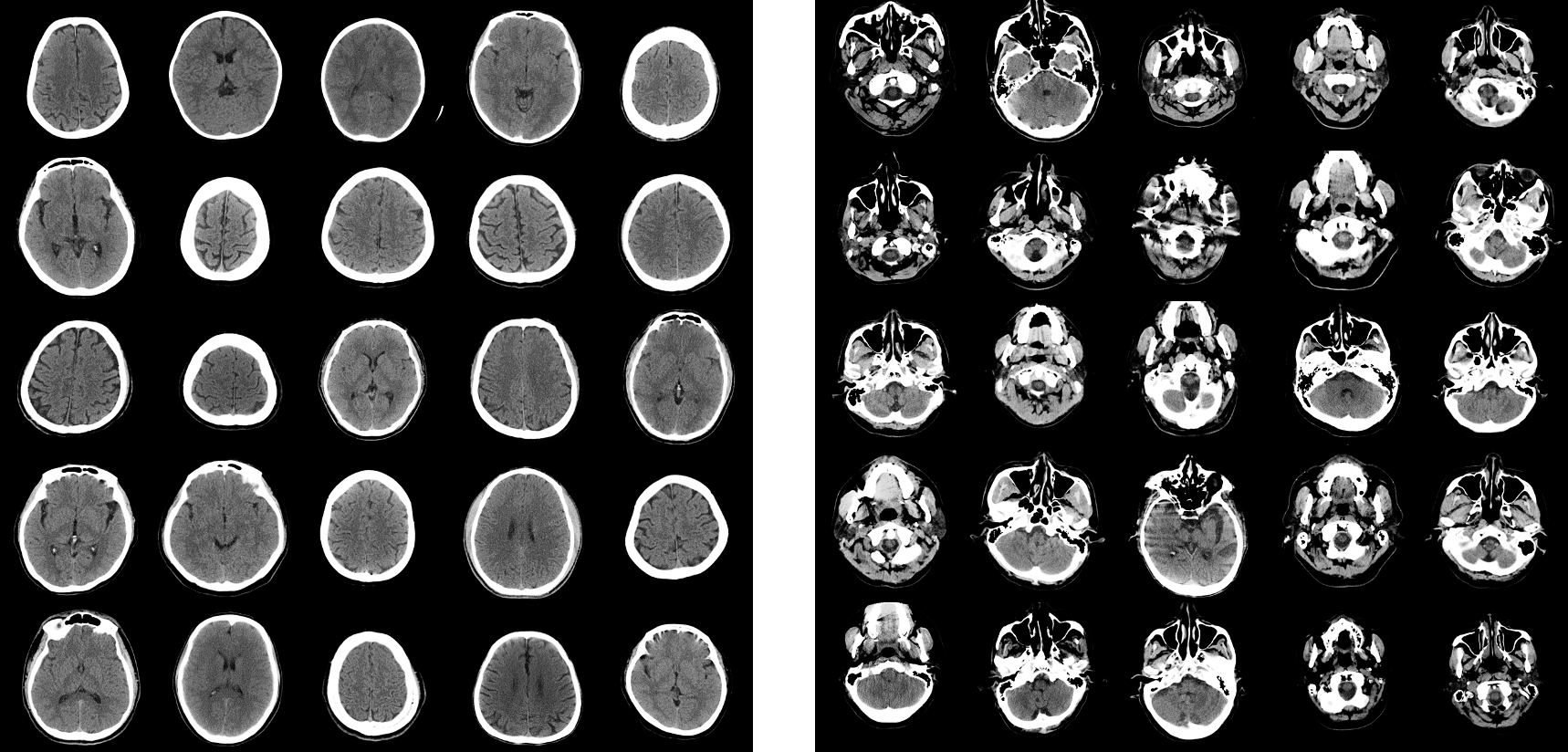}}
\caption{(Left) Examples of supratentorial part of the brain CT images. (Right) Examples of infratentorial part of the brain CT images.}
\label{supinf}
\end{center}
\vskip -0.2in
\end{figure}

As shown in Figure (\ref{brainct_ex}), blank, empty images are also generated, due to the fact that original CT images contain blank, empty images above the head.\\
We experimented 2 schemes: (1) Calculating precision, recall, density and coverage (PRDC), and fidelity, diversity using barcode between real and synthetic images. (2) One licensed medical doctor manually split 1,000 images containing the superior part and 1,000 images containing the inferior part of the brain in real brain CT images. Criteria of splitting superior and inferior brain CT images was eye level, implying superior part of the brain CT was above the eyes, and inferior part of the brain was below the eyes. We call superior to the eye as \textbf{supratentorial} and inferior to the eye as \textbf{infratentorial}. In almost all cases, supratentorial images contain skull and brain parenchyma, however infratentorial images contain maxilla, mandible. Example images of supratentorial and infratentorial part of the brain are shown in Figure (\ref{supinf}).

\begin{figure}[h]
\vskip 0.2in
\begin{center}
\centerline{\includegraphics[height=50mm]{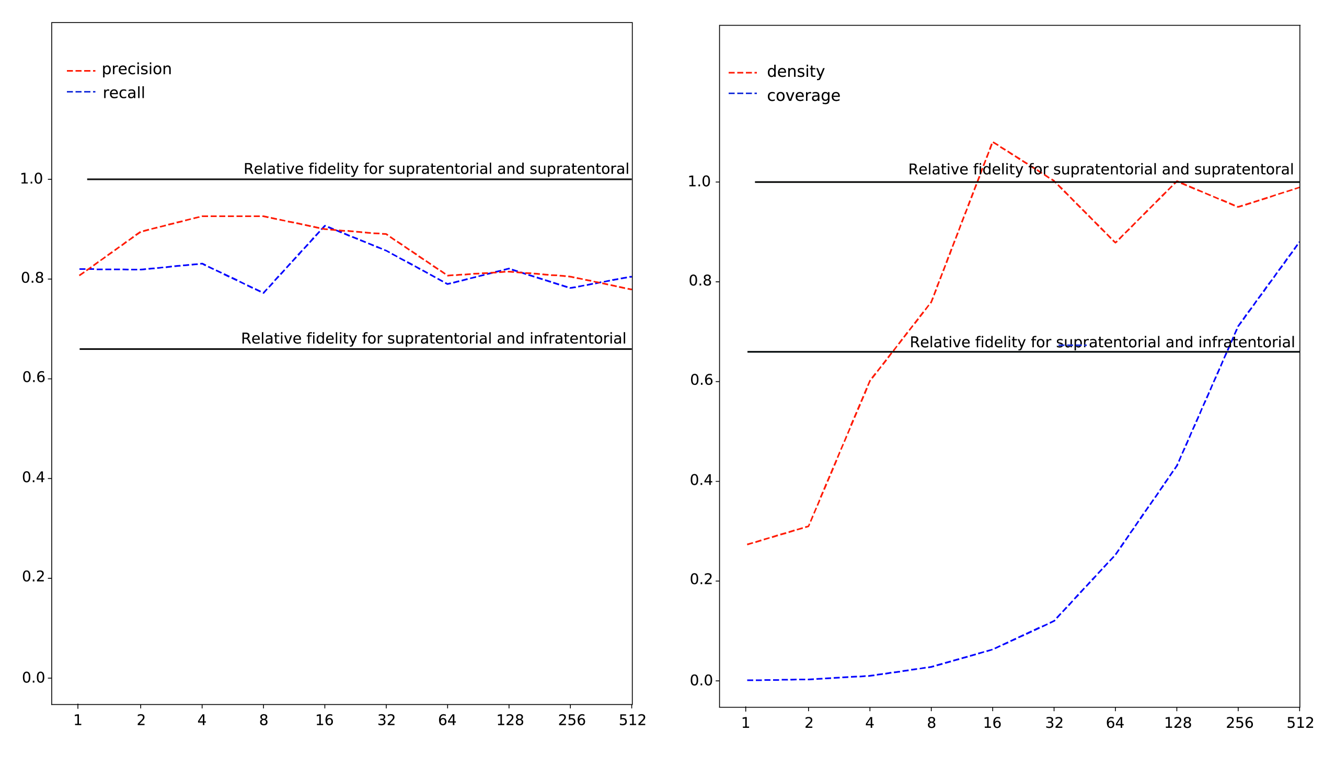}}
\caption{(Left) Precision and recall for randomly swapped images. (Right) Density and coverage for randomly swapped images. $x$ axis implies the number of randomly swapped images.}
\label{stress_test}
\end{center}
\vskip -0.2in
\end{figure}

\begin{figure}[h]
\vskip 0.2in
\begin{center}
\centerline{\includegraphics[width=70mm]{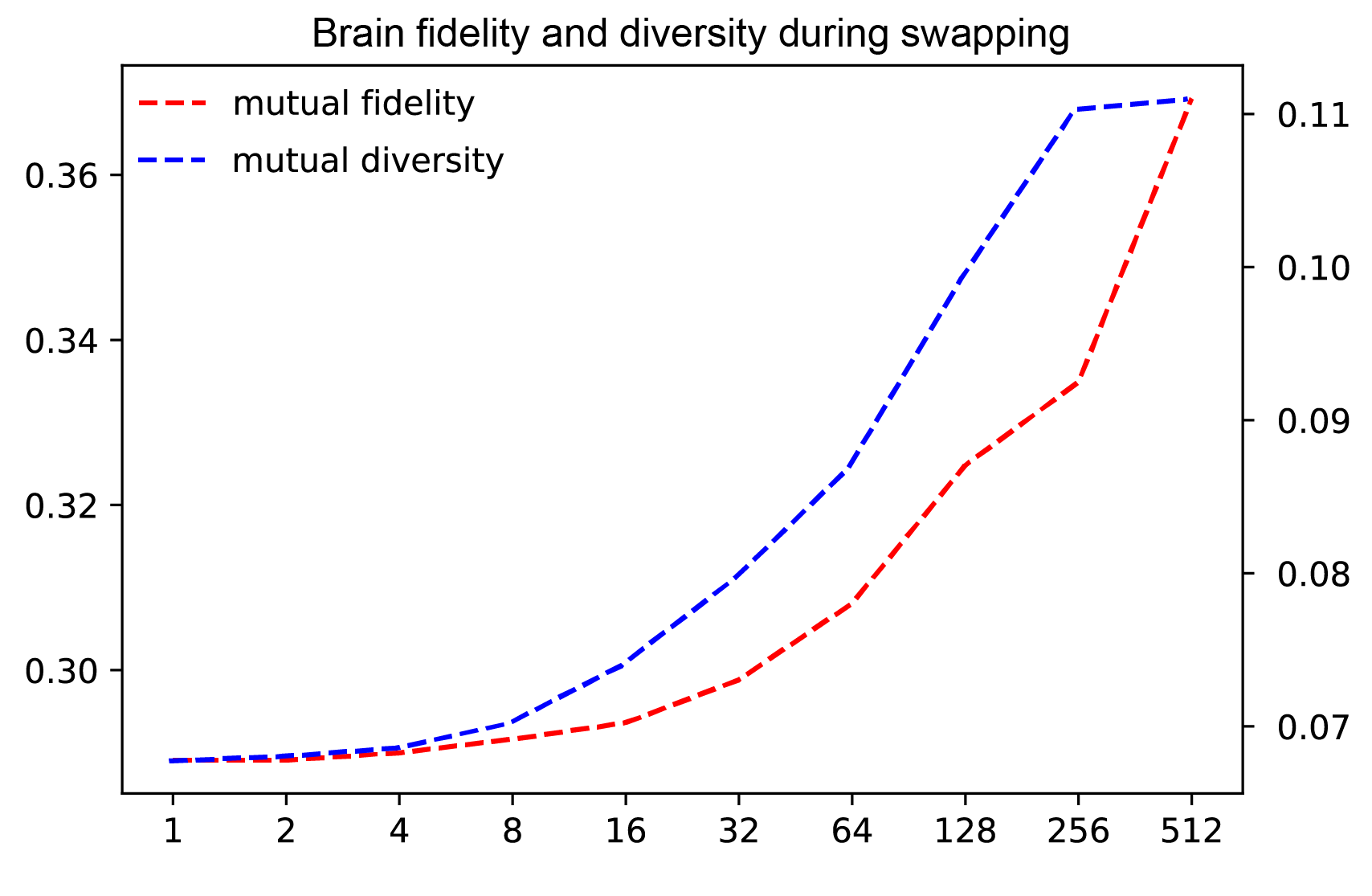}}
\caption{Relative fidelity and relative diversity of randomly swapped images. $x$ axis implies the number of randomly swapped images. For $y$ axis, left axis ranging from 0.30 to 0.36 indicates mutual fidelity and right axis ranging from 0.07 to 0.11 indicates mutual diversity.}
\label{brain_stress_test}
\end{center}
\vskip -0.2in
\end{figure}

\begin{table*}[b!]
\caption{Evaluation of various metrics on 1,000 supratentorial (supra) and 1,000 infratentorial (infra) real brain CT dataset. $^{\ast}$ implies our proposed metric. Here, there were no real and fake images. Therefore, relative fidelity and relative diversity were not mentioned.}
\label{table_supinf}
\vskip 0.15in
\begin{center}
\begin{small}
\begin{sc}
\begin{tabular}{l|ccc}
 & Supra and Infra & Supra and Supra & Infra and Infra\\
 \hline
Precision & 0.000 &1.000 & 1.000\\
Recall & 0.003 & 1.000&1.000\\
Density & 0.000 &0.977 &0.983\\
Coverage & 0.000 &1.000 &1.000\\
Fidelity$^{\ast}$ & 0.289 & 0.382 & 0.343 \\
Diversity$^{\ast}$ & 0.067 & 0.101 & 0.081 \\
\end{tabular}
\end{sc}
\end{small}
\end{center}
\vskip -0.1in
\end{table*}

Result of supretentorial and infratentorial images are shown in Table (\ref{table_supinf}).\\

We also performed a stress test for supratentorial and infratentorial brain CT images. To perform a stress test, from sampled 1,000 supratentorial and infratentorial brain CT images, we randomly swapped images between supratentorial and infratentorial images. In PRDC paper, the authors performed experiments on random samples from normal distribution. The robustness of density and coverage is considered to come from $k$-nearest neighborhood method. However, in this paper, we perfrom stress test on real dataset, which is more close to real setting as well as inappropriateness of normal distribution as discussed in section \ref{section31}. For $0\leq n\leq 9$, we randomly swapped $2^n$ images respectively, and calculated PRDC, and fidelity. The results are shown in Figure (\ref{stress_test}) and Figure (\ref{brain_stress_test}). As shown in the figure, precision and recall does not catch outliers at all. These two metrics fluctuated, and had no tendency of detecting outliers even if outlier probability increased from 0.1\% ($n=0$) to 51.2\% ($n=9$). Compared to these two metrics, density and coverage had tendency to increase when outliers are given. However, density was too sensitive to outliers - which seemed to increase significantly at 0.4\% ($n=2$). Coverage was the most reasonable metric in detecting outliers. However, coverage increased significantly at rate 6.4\% ($n=6$) , which was not robust to outliers. With our proposed method, fidelity was the most robust metric as well as had an ability to detect outliers with a reasonable performance. Using barcode, fidelity had robustness up to 6.4\%, and began to detect outliers from 6.4\%. Furthermore, as ratio ($n$) increased, the distributions of two datasets became similar. Therefore diversity was decreased. In this experiment, there was no fake distribution, and as both datasets were real, we divided diversity of supratentorial brain CT images to diversity of infratentorial brain CT images with unswapped settings.\\

Finally, we had experimented inserting abdomen CT images among brain CT images to observe how metrics act to outliers. This is discussed in Appendix \ref{appendixb}.

\section{Discussion}\label{discussion}

Our proposed barcode has several benefits: (1) Barcode does not require hyperparameters except dimension reduction and graph plotting. (2) Barcode does require strong assumptions except for $L^2$ distance. The only critical assumption of barcode is that similar images locate in a near position in embedded space by the Inception network. (3) It not only measures fidelity, but also measures diversity. Furthermore, as we had divided diversity into two terms - intrinsic and extrinsic diversities - we can measure detailed diversities in special occasions. Furthermore, as discussed in section \ref{experiments}, our metric outperforms traditional methods such as precision and recall, density and coverage in intuitive manner. Furthermore, as shown in tables, our metric is more stable than other traditional methods.\\
As discussed in chapter \ref{section31}, we had mathematically proved that normal assumptions are not valid in embedded spaces, and validated this theoretical results with experiments using barcode plot (Figure (\ref{normal_barcode})). Therefore metrics such as FID should be modified with assumption-free methods, such as our proposed metric, barcode. Furthermore, metrics such as PRDC use hyperparameters derived by $k$NN methods. Our proposed metric does not require hyperparameters even more, though user-selective hyperparameters are implemented in codes. Note that our experiments did not use any hyperparameter, such as reducing explanablity, outlier removal.\\
In addition, the abstraction of our proposed metric, barcode can be performed. Our method can be applied to any metrizable topological space, equipped with sound metrics that can express the topology of embedded vectors.\\
Our extensive experiments from evaluation of GAN-generated images to swapping, and even inserting abdomen images into brain images show barcode works better than previously proposed metrics, such as PRDC or FID. \\

In conclusion, we had proposed a novel method motivated by topological data analysis, namely barcode. With our proposed method, one can measure diversity and fidelity of generation models quantitatively, and our experiments showed our metric outperforms traditional methods.

\printbibliography

@article{goodfellow2014generative,
  title={Generative adversarial nets},
  author={Goodfellow, Ian and Pouget-Abadie, Jean and Mirza, Mehdi and Xu, Bing and Warde-Farley, David and Ozair, Sherjil and Courville, Aaron and Bengio, Yoshua},
  journal={Advances in neural information processing systems},
  volume={27},
  pages={2672--2680},
  year={2014}
}

@inproceedings{Choi2018StarGANUG,
  title={Stargan: Unified generative adversarial networks for multi-domain image-to-image translation},
  author={Choi, Yunjey and Choi, Minje and Kim, Munyoung and Ha, Jung-Woo and Kim, Sunghun and Choo, Jaegul},
  booktitle={Proceedings of the IEEE conference on computer vision and pattern recognition},
  pages={8789--8797},
  year={2018}
}

@inproceedings{Zhu2017UnpairedIT,
  title={Unpaired image-to-image translation using cycle-consistent adversarial networks},
  author={Zhu, Jun-Yan and Park, Taesung and Isola, Phillip and Efros, Alexei A},
  booktitle={Proceedings of the IEEE international conference on computer vision},
  pages={2223--2232},
  year={2017}
}

@inproceedings{Choi2020StarGANVD,
  title={Stargan v2: Diverse image synthesis for multiple domains},
  author={Choi, Yunjey and Uh, Youngjung and Yoo, Jaejun and Ha, Jung-Woo},
  booktitle={Proceedings of the IEEE/CVF Conference on Computer Vision and Pattern Recognition},
  pages={8188--8197},
  year={2020}
}

@article{brock2018large,
  title={Large scale GAN training for high fidelity natural image synthesis},
  author={Brock, Andrew and Donahue, Jeff and Simonyan, Karen},
  journal={arXiv preprint arXiv:1809.11096},
  year={2018}
}

@article{radford2015unsupervised,
  title={Unsupervised representation learning with deep convolutional generative adversarial networks},
  author={Radford, Alec and Metz, Luke and Chintala, Soumith},
  journal={arXiv preprint arXiv:1511.06434},
  year={2015}
}

@article{karras2017progressive,
  title={Progressive growing of gans for improved quality, stability, and variation},
  author={Karras, Tero and Aila, Timo and Laine, Samuli and Lehtinen, Jaakko},
  journal={arXiv preprint arXiv:1710.10196},
  year={2017}
}

@inproceedings{karras2019style,
  title={A style-based generator architecture for generative adversarial networks},
  author={Karras, Tero and Laine, Samuli and Aila, Timo},
  booktitle={Proceedings of the IEEE conference on computer vision and pattern recognition},
  pages={4401--4410},
  year={2019}
}

@inproceedings{karras2020analyzing,
  title={Analyzing and improving the image quality of stylegan},
  author={Karras, Tero and Laine, Samuli and Aittala, Miika and Hellsten, Janne and Lehtinen, Jaakko and Aila, Timo},
  booktitle={Proceedings of the IEEE/CVF Conference on Computer Vision and Pattern Recognition},
  pages={8110--8119},
  year={2020}
}

@article{karras2020training,
  title={Training generative adversarial networks with limited data},
  author={Karras, Tero and Aittala, Miika and Hellsten, Janne and Laine, Samuli and Lehtinen, Jaakko and Aila, Timo},
  journal={arXiv preprint arXiv:2006.06676},
  year={2020}
}

@article{salimans2016improved,
  title={Improved techniques for training gans},
  author={Salimans, Tim and Goodfellow, Ian and Zaremba, Wojciech and Cheung, Vicki and Radford, Alec and Chen, Xi},
  journal={arXiv preprint arXiv:1606.03498},
  year={2016}
}

@inproceedings{szegedy2016rethinking,
  title={Rethinking the inception architecture for computer vision},
  author={Szegedy, Christian and Vanhoucke, Vincent and Ioffe, Sergey and Shlens, Jon and Wojna, Zbigniew},
  booktitle={Proceedings of the IEEE conference on computer vision and pattern recognition},
  pages={2818--2826},
  year={2016}
}

@inproceedings{heusel2017gans,
  title={Gans trained by a two time-scale update rule converge to a local nash equilibrium},
  author={Heusel, Martin and Ramsauer, Hubert and Unterthiner, Thomas and Nessler, Bernhard and Hochreiter, Sepp},
  booktitle={Advances in neural information processing systems},
  pages={6626--6637},
  year={2017}
}

@inproceedings{sajjadi2018assessing,
  title={Assessing generative models via precision and recall},
  author={Sajjadi, Mehdi SM and Bachem, Olivier and Lucic, Mario and Bousquet, Olivier and Gelly, Sylvain},
  booktitle={Advances in Neural Information Processing Systems},
  pages={5228--5237},
  year={2018}
}

@inproceedings{kynkaanniemi2019improved,
  title={Improved precision and recall metric for assessing generative models},
  author={Kynk{\"a}{\"a}nniemi, Tuomas and Karras, Tero and Laine, Samuli and Lehtinen, Jaakko and Aila, Timo},
  booktitle={Advances in Neural Information Processing Systems},
  pages={3927--3936},
  year={2019}
}

@article{naeem2020reliable,
  title={Reliable Fidelity and Diversity Metrics for Generative Models},
  author={Naeem, Muhammad Ferjad and Oh, Seong Joon and Uh, Youngjung and Choi, Yunjey and Yoo, Jaejun},
  journal={arXiv preprint arXiv:2002.09797},
  year={2020}
}

@article{carlsson2005persistence,
  title={Persistence barcodes for shapes},
  author={Carlsson, Gunnar and Zomorodian, Afra and Collins, Anne and Guibas, Leonidas J},
  journal={International Journal of Shape Modeling},
  volume={11},
  number={02},
  pages={149--187},
  year={2005},
  publisher={World Scientific}
}

@article{blum2016foundations,
  title={Foundations of data science},
  author={Blum, Avrim and Hopcroft, John and Kannan, Ravindran},
  journal={Vorabversion eines Lehrbuchs},
  volume={5},
  year={2016}
}

\appendix

\section{Appendix}\label{appendixex}
\begin{figure}[ht]
\vskip 0.2in
\begin{center}
\centerline{\includegraphics[width=80mm]{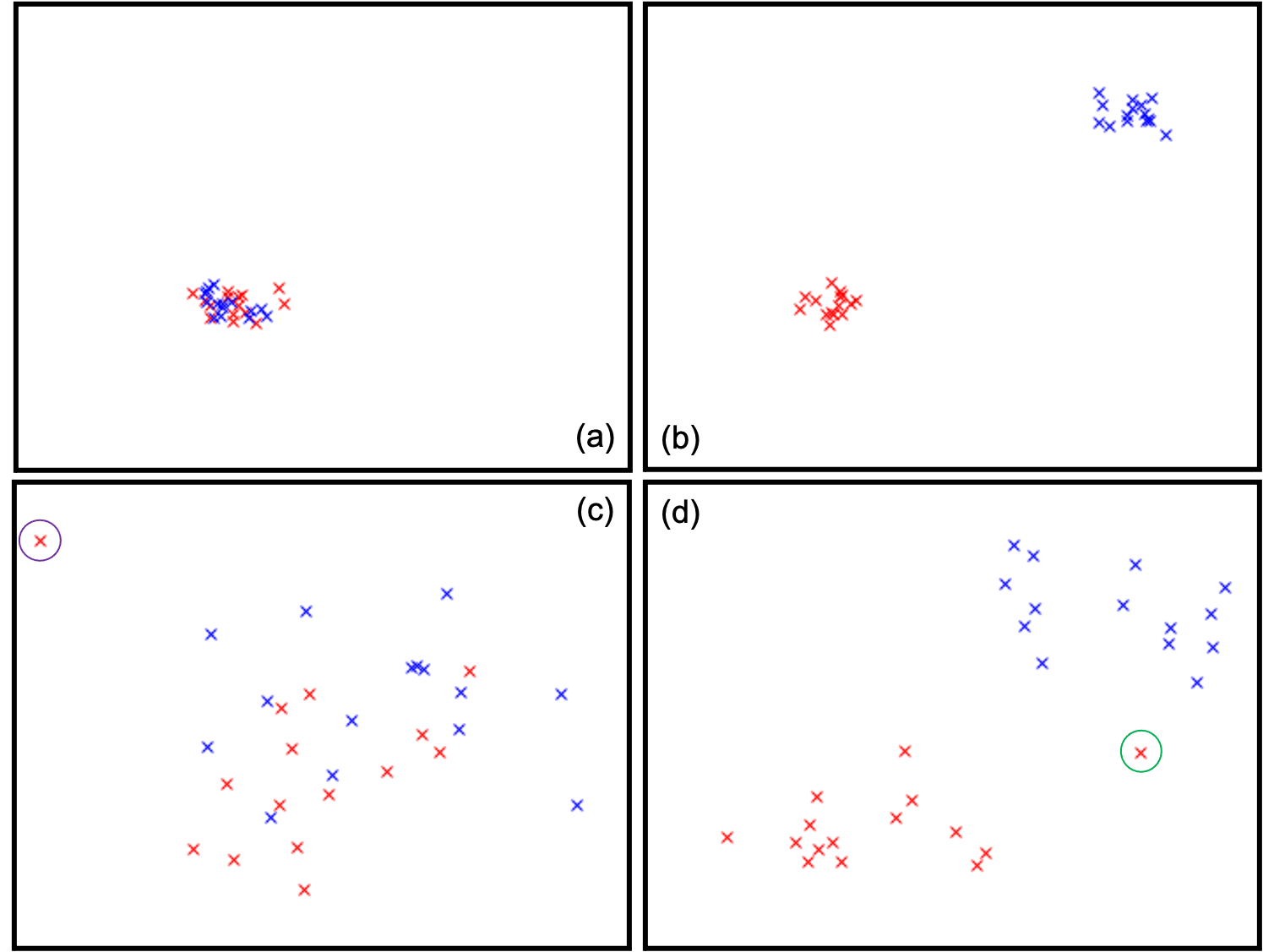}}
\caption{Four cases of fidelity and diversity. (a) High fidelity, low diversity. (b) Low fidelity, low diversity. (c) High fidelity, high diversity. Purple circle denotes outlier. (d) Low fidelity, high diversity. The green circle denotes an outlier.}
\label{fig1}
\end{center}
\vskip -0.2in
\end{figure}

\begin{figure}[ht]
\vskip 0.2in
\begin{center}
\centerline{\includegraphics[width=60mm]{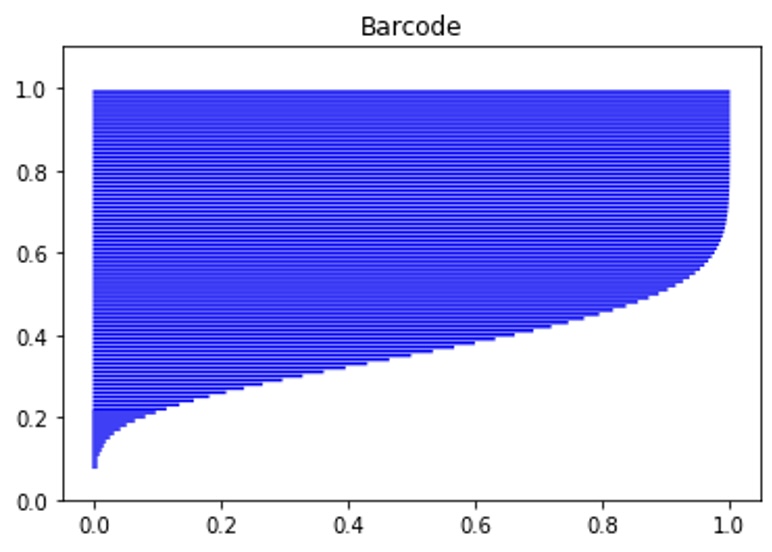}}
\caption{Example image of barcode plot.}
\label{barcode_ex}
\end{center}
\vskip -0.2in
\end{figure}

\begin{figure}[ht]
\vskip 0.2in
\begin{center}
\centerline{\includegraphics[width=100mm]{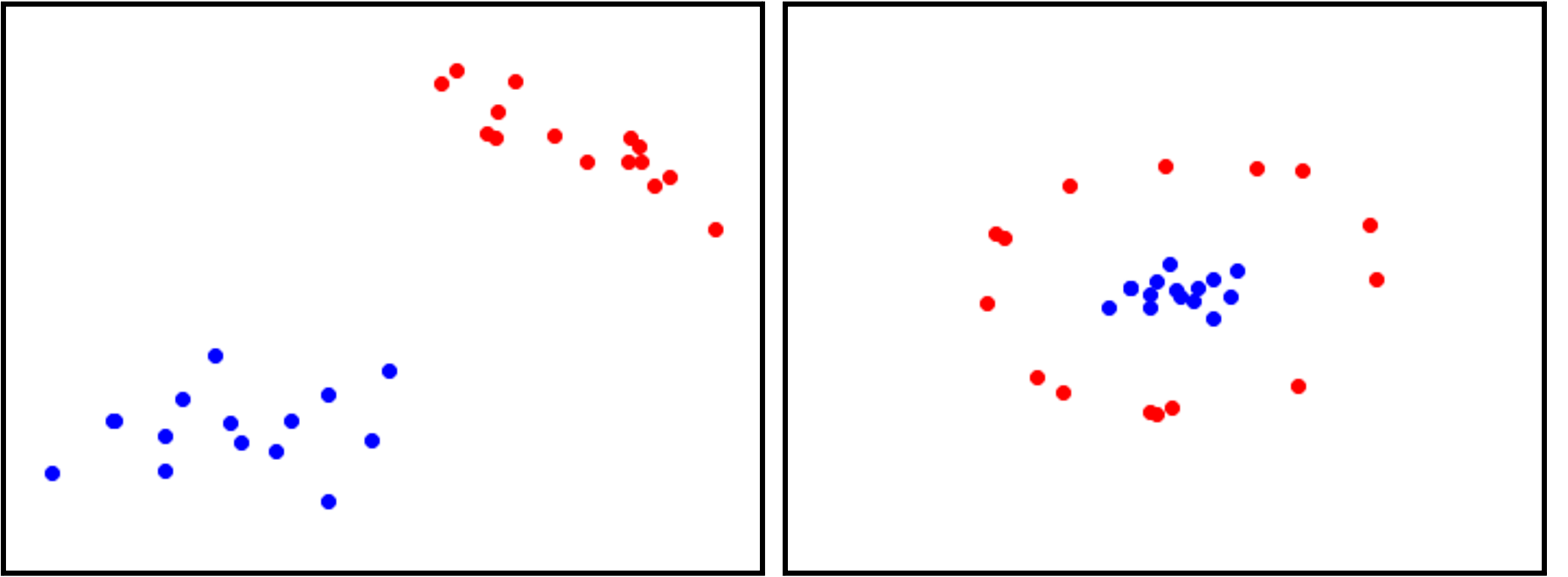}}
\caption{Two different cases that have similar fidelity (left: 0.84, right: 0.87) and extrinsic diversity (left: 0.13, right: 0.14). Red dots in the left figure have small relative diversity (0.53) while red dots in the right figure have large relative diversity (0.62). Due to the fact that relative diversity is calculated from two intrinsic diversities ($\delta(P,P)$ and $\delta(Q,Q)$), we calculated relative diversity of two situations.}
\label{fig2}
\end{center}
\vskip -0.2in
\end{figure}

In this section we provide example figures for four cases ($\{$high fidelity, low fidelity$\}\times\{$high diversity, low diversity$\}$) and example figure of differentiating extrinsic and relative diversity. See Figure (\ref{fig1}) to (\ref{fig2}).

\section{Appendix}\label{appendix}

\begin{figure}[ht]
\vskip 0.2in
\begin{center}
\centerline{\includegraphics[width=120mm]{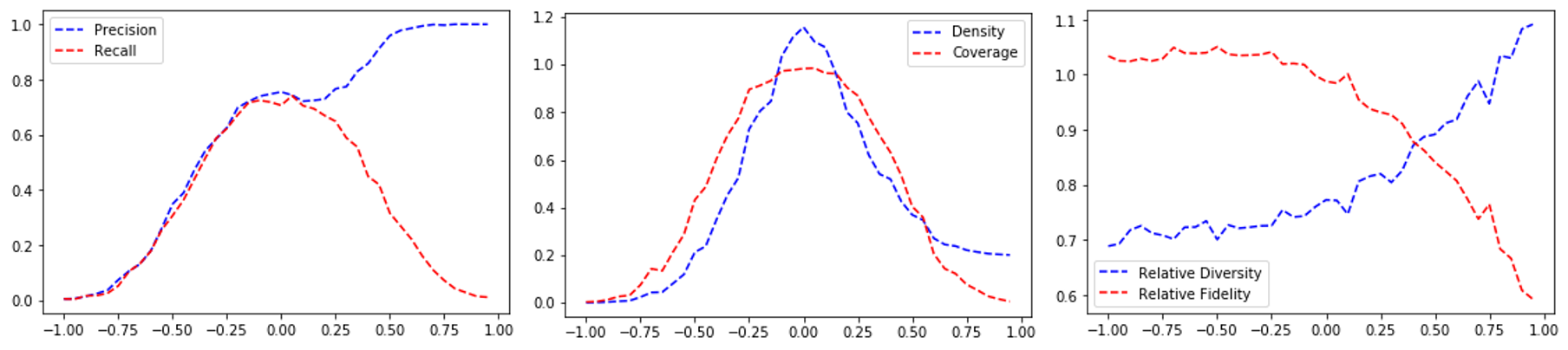}}
\caption{Experiments to analyze robustness on outliers. (Left) Precision and recall when one outlier $(3,\cdots,3)$ was added on 999 normal samples. (Middle) Density and coverage when one outlier $(3,\cdots,3)$ was added on 999 normal samples. (Right) Fidelity and diversity when one outlier $(3,\cdots,3)$ was added on 999 normal samples. In PRDC experiments, $k$ was set to be 5.}
\label{outlier_robustness}
\end{center}
\vskip -0.2in
\end{figure}

In PRDC paper, the authors had performed experiments on 64-dimensional normal dataset to scrutinize robustness of their proposed algorithm. Let $X$ follow 64-dimensional normal distribution with variance 1 at each direction, and have no covariance between dimensions. To see how robust density and coverage on outliers, we had sampled 999 samples from this 64-dimensional normal distribution, and added one 64-dimensional vector, with all elements are 3. That is, 999 samples follow normal distribution, and one sample is $(3,\cdots,3)$. The results on PRDC, fidelity and diversity based on barcode is shown in Figure (\ref{outlier_robustness}).\\
In one glimpse, our fidelity and diversity does not seem robust to outlier. However, in this section, we will discuss that one 64-dimensional outlier vector $(3,\cdots,3)$ is too harsh condition to analyze robustness on outliers.\\
As discussed in section \ref{section31}, GAT says that most probability of high-diemsional normal distribution lies in some annulus. If we formulate GAT with $d=64$ setting, we get following theorem \cite{blum2016foundations}:
\begin{thm}\label{GAT64}
For a 64-dimensional spherical Gaussian with unit variance in each dimension for any $\beta\leq8$, all but at most $3e^{-c\beta^2}$ of the probability mass lies within the annulus $8-\beta\leq|x|\leq8+\beta$, where $c$ is a fixed positive constant.
\end{thm}

\begin{figure}[ht]
\vskip 0.2in
\begin{center}
\centerline{\includegraphics[width=50mm]{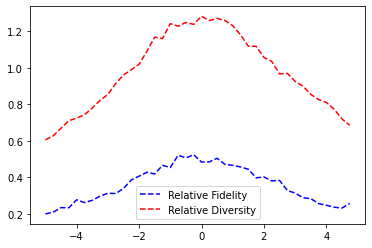}}
\caption{Barcode values according to mean ranging from $-5$ to $5$. Outlier probability was set to be $0.001$, with position `out', impliying 0.1\% of the largest distances are considered to be outliers.}
\label{stress_test_outlier}
\end{center}
\vskip -0.2in
\end{figure}

This statement says that most probability lies in annulus with at most diameter 16. However, outlier in the previous experiment have distance $\sqrt{\sum_{i=1}^{64}3^2}=24$. To detour this problem, we used outlier removal algorithm in section \ref{section_outlier_removal}. We set outlier probability with 0.001 and outlier position to be `out', which means with 0.1\% ratio of the largest distances are outliers and experimented above setting. The result is shown in Figure (\ref{stress_test_outlier})

\section{Appendix}\label{appendixc}
\subsection{CelebA-HQ dataset}
We performed the experiment on CelebA-HQ dataset. We calculated various metrics on 50,000 real and synthetic images. We used pre-trained PGGAN, StyleGAN models to evaluate model performance. For experiments, the results are shown in Table (\ref{celeba_results})

\begin{table*}[t]
\caption{Evaluation of various metrics on 50,000 real and 50,000 synthetic CelebA-HQ dataset on PGGAN and StyleGAN pre-trained networks. $^{\ast}$ implies our proposed metric.}
\label{celeba_results}
\vskip 0.15in
\begin{center}
\begin{small}
\begin{sc}
\begin{tabular}{l|c|c|c|c}

 && \multicolumn{1}{p{2cm}}{\centering Real \\ and \\ Synthetic} & \multicolumn{1}{|p{2cm}}{\centering Real \\ and \\ Real} & \multicolumn{1}{|p{2cm}}{\centering Synthetic \\ and \\ Synthetic}\\
\hline
PGGAN &Precision &0.146&&\\
&Recall& 0.011& &\\
& Density & 0.069 &&\\
& Coverage & 0.052&&\\
& Fidelity$^{\ast}$ & 0.519$^{a}$&0.489$^b$&0.510\\
& Diversity$^{\ast}$ &0.071$^{c}$& 0.088$^{d}$& 0.085$^{e}$\\
& Relative fidelity$^{\ast}$ & 1.061$^{a/b}$&& \\
& Relative diversity$^{\ast}$ &0.826$^{c/(\sqrt{d}\sqrt{e})}$ && \\
\hline
StyleGAN &Precision &0.154& &\\
&Recall&0.011 & &\\
& Density & 0.073 & &\\
& Coverage & 0.057& &\\
& Fidelity$^{\ast}$ & 0.484$^{a}$&0.489$^b$&0.507\\
& Diversity$^{\ast}$ &0.077$^{c}$& 0.088$^{d}$& 0.087$^{e}$\\
& Relative fidelity$^{\ast}$ & 0.988$^{a/b}$&& \\
& Relative diversity$^{\ast}$ &0.878$^{c/(\sqrt{d}\sqrt{e})}$ && \\

\end{tabular}
\end{sc}
\end{small}
\end{center}
\vskip -0.1in
\end{table*}

\subsection{FFHQ dataset}
For FFHQ dataset, we randomly sampled 50,000 real images and 50,000 synthetic images for StyleGAN, StyleGAN2, StyleGAN-Ada models with various metrics. The results are shown in Table (\ref{FFHQ_table})

\begin{table*}[t]
\caption{Evaluation of various metrics on 50,000 real and 50,000 synthetic FFHQ dataset on StyleGAN and StyleGAN2, StyleGAN-Ada pre-trained networks. $^{\ast}$ implies our proposed metric.}
\label{FFHQ_table}
\vskip 0.15in
\begin{center}
\begin{small}
\begin{sc}
\begin{tabular}{l|c|c|c|c}

 && \multicolumn{1}{p{2cm}}{\centering Real \\ and \\ Synthetic} & \multicolumn{1}{|p{2cm}}{\centering Real \\ and \\ Real} & \multicolumn{1}{|p{2cm}}{\centering Synthetic \\ and \\ Synthetic}\\
\hline
StyleGAN &Precision &0.704&&\\
&Recall& 0.419& &\\
& Density & 0.152 &&\\
& Coverage & 0.811&&\\
& Fidelity$^{\ast}$ & 0.556$^{a}$&0.545$^b$&0.478\\
& Diversity$^{\ast}$ &0.062$^{c}$& 0.063$^{d}$& 0.073$^{e}$\\
& Relative fidelity$^{\ast}$ & 1.020$^{a/b}$&& \\
& Relative diversity$^{\ast}$ &0.908$^{c/(\sqrt{d}\sqrt{e})}$ && \\
\hline
StyleGAN2 &Precision &0.681& &\\
&Recall&0.519& &\\
& Density & 0.750 & &\\
& Coverage & 0.807& &\\
& Fidelity$^{\ast}$ & 0.543$^{a}$&0.545$^b$&0.527\\
& Diversity$^{\ast}$ &0.063$^{c}$& 0.063$^{d}$& 0.066$^{e}$\\
& Relative fidelity$^{\ast}$ & 0.996$^{a/b}$&& \\
& Relative diversity$^{\ast}$ &0.977$^{c/(\sqrt{d}\sqrt{e})}$ && \\
\hline
StyleGAN2-Ada &Precision &0.679& &\\
&Recall&0.504 & &\\
& Density & 0.730 & &\\
& Coverage & 0.812& &\\
& Fidelity$^{\ast}$ & 0.555$^{a}$&0.545$^b$&0.518\\
& Diversity$^{\ast}$ &0.062$^{c}$& 0.063$^{d}$& 0.067$^{e}$\\
& Relative fidelity$^{\ast}$ & 1.018$^{a/b}$&& \\
& Relative diversity$^{\ast}$ &0.946$^{c/(\sqrt{d}\sqrt{e})}$ && \\

\end{tabular}
\end{sc}
\end{small}
\end{center}
\vskip -0.1in
\end{table*}

\section{Appendix}\label{appendixb}
\begin{figure}[h!]
\vskip 0.2in
\begin{center}
\centerline{\includegraphics[width=130mm]{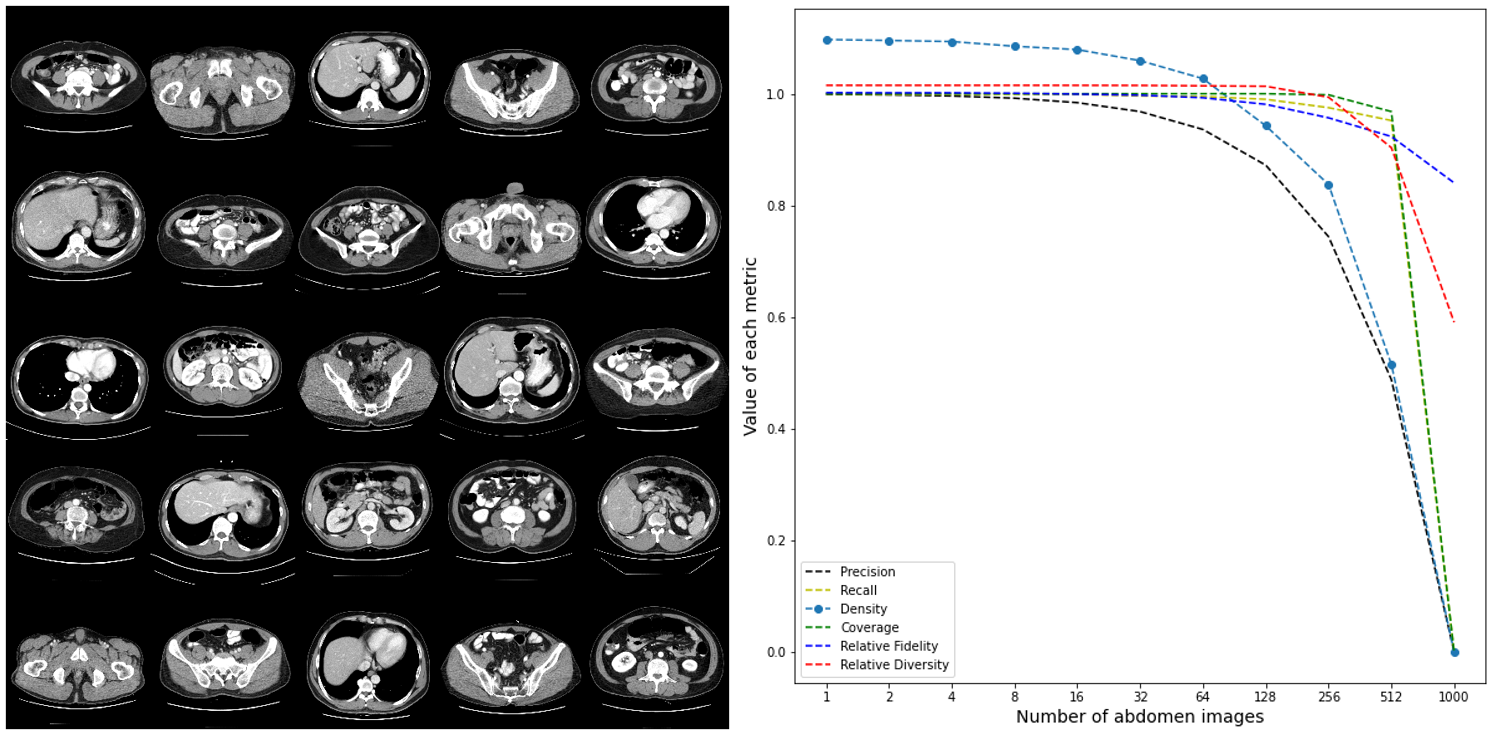}}
\caption{Inserting abdomen CT to 1,000 brain CT images. (Left) Example 25 images of abdomen CT. (Right) Total number of images are always 1,000. $x$ axis denote number of abdomen images, $y$ axis denote value of each metric. }
\label{abdomen_ex_and_graph}
\end{center}
\vskip -0.2in
\end{figure}

First, we had sampled 1,000 real brain CT images randomly. Then, for $i$ in range $0$ to $10$, we replaced $2^i$ brain CT images to abdomen CT images(As we only sampled 1,000 images, $2^10$ was considered as 1,000). The result is shown in Figure (\ref{abdomen_ex_and_graph}). As the result shows, metrics such as precision, recall, density, coverage converge to zero as n increases. However, our proposed barcode metrics, which is fidelity and diversity does not converge to zero. This is basically barcode metrics are based on distances, which cannot be zero unless all distances are zeros. More specifically, density and precision are not that robust compare to our metrics.

\end{document}